\documentclass[10pt,twocolumn,letterpaper]{article}

\usepackage[pagenumbers]{wacv} 

\usepackage{graphicx}
\usepackage{booktabs}

\usepackage[pagebackref,breaklinks,colorlinks]{hyperref}

\usepackage{algpseudocode}
\usepackage{algorithm}
\usepackage{amsmath}

\usepackage{multirow}
\usepackage{multicol}
\usepackage{makecell}
\usepackage{diagbox}
\usepackage{tabularx}
\usepackage{amssymb}
\usepackage[dvipsnames]{xcolor}
\usepackage[pagebackref,breaklinks,colorlinks]{hyperref}
\usepackage{pifont}

\usepackage{verbatim}
\usepackage{titlesec}
\usepackage{slashbox}

\titleformat{\section}
  {\normalfont\large\bfseries} 
  {\thesection.} 
  {1em} 
  {} 
  [] 

\titleformat{\subsection}
  {\normalfont\large\bfseries} 
  {\thesubsection.} 
  {1em} 
  {} 
  [] 

\usepackage[capitalize]{cleveref}
\crefname{section}{Sec.}{Secs.}
\Crefname{section}{Section}{Sections}
\Crefname{table}{Table}{Tables}
\crefname{table}{Tab.}{Tabs.}

\def\confName{WACV}
\def\confYear{2025}

\begin{document}

\title{Label-Augmented Dataset Distillation}


\author{
Seoungyoon Kang$^{1,2}$\footnotemark[1], Youngsun Lim$^{2}$\footnotemark[1], Hyunjung Shim$^{2}$ \\
$^{1}$Yonsei University, 
$^{2}$KAIST AI\\
{\tt\small sy.kang@yonsei.ac.kr, \{youngsun\_ai, kateshim\}@kaist.ac.kr}
}
\renewcommand{\thefootnote}{\fnsymbol{footnote}}
\maketitle
\footnotetext[1]{Equal contribution}

\begin{abstract}
Traditional dataset distillation primarily focuses on image representation while often overlooking the important role of labels. In this study, we introduce Label-Augmented Dataset Distillation (LADD), a new dataset distillation framework enhancing dataset distillation with label augmentations. LADD sub-samples each synthetic image, generating additional dense labels to capture rich semantics. These dense labels require only a 2.5\% increase in storage (ImageNet subsets) with significant performance benefits, providing strong learning signals. Our label-generation strategy can complement existing dataset distillation methods and significantly enhance their training efficiency and performance. Experimental results demonstrate that LADD outperforms existing methods in terms of computational overhead and accuracy. With three high-performance dataset distillation algorithms, LADD achieves remarkable gains by an average of 14.9\% in accuracy. 
Furthermore, the effectiveness of our method is proven across various datasets, distillation hyperparameters, and algorithms. Finally, our method improves the cross-architecture robustness of the distilled dataset, which is important in the application scenario.
\end{abstract}

\section{Introduction}
\label{sec:intro}

\begin{figure*}[t]
  \centering
  \includegraphics[width=0.89\textwidth]{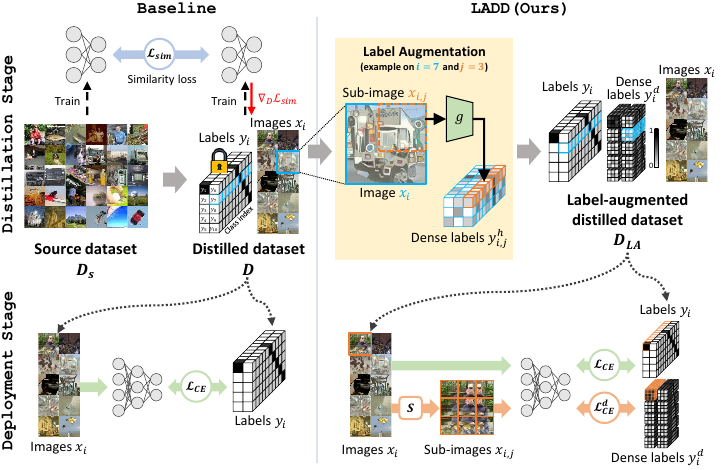}
  \vspace{-0.1in}
  \caption{
        \textbf{Overview of LADD.} Once the distilled dataset $D$ is synthesized by baseline, LADD initiates label augmentation. It divides each image in $D$ into $N \times N$ sub-images, as illustrated in Fig. 1 ($N=3$). Then, $N^2$ soft labels are computed using the labeler $g$ to produce the dense label. Label augmented distilled dataset $D_{LA}$ consists of images, labels, and dense labels; it is utilized in the deployment stage to train the evaluation model.}
  \label{fig:overview}
  \vspace{-0.17in}
\end{figure*}

Dataset distillation, also called dataset condensation, creates a small synthetic training set to reduce training costs. The synthesized dataset enables faster training while maintaining a performance comparable to that achieved with the source dataset.
For example, FrePo~\cite{ref_frepo_zhou2022dataset} attained 93\% of full dataset training performance using merely one image per class in MNIST~\cite{ref_mnist_deng2012mnist}. Dataset distillation can be applied in various fields. These include privacy-free training data generation (e.g., federated learning~\cite{ref_federate_goetz2020federated,ref_secdd_sucholutsky2020secdd,ref_oneshot_zhou2020distilled}, medical image computing~\cite{ref_xray_li2020soft,ref_secdd_sucholutsky2020secdd}), fast training (e.g., network architecture search~\cite{ref_gradmat_zhao2021dataset,ref_distmat_zhao2023dataset,ref_dsa_zhao2021dataset}), or compact training data generation (e.g., continual learning~\cite{ref_gradmat_zhao2021dataset,ref_distmat_zhao2023dataset,ref_dsa_zhao2021dataset}).

The efficacy of distilled datasets is typically evaluated based on the test accuracy achieved by models trained by these datasets. The distilled dataset must maximally encapsulate essential information of the source dataset within a limited number of synthetic samples. Prior research~\cite{ref_dd_wang2018dataset,ref_distmat_zhao2023dataset,ref_gradmat_zhao2021dataset,ref_mtt_cazenavette2022dataset,ref_krr_nguyen2021dataset,ref_implicit_vicol2022implicit,ref_convex_loo2023dataset} has refined the optimization objective within the bi-loop nested meta-learning framework for dataset synthesis. Some methods have further explored optimization spaces beyond image~\cite{ref_glad_cazenavette2023generalizing,ref_latentdd_duan2023dataset} and efficient ways to utilize pixel-space~\cite{ref_idc_kim2022dataset}. Additionally, several approaches~\cite{ref_frepo_zhou2022dataset, ref_tesla_cui2023scaling, ref_dim_wang2023dim} develop algorithms to reduce the computational cost induced by the bi-loop optimization. However, these efforts mostly focus on data representation in images, overlooking the important roles of labels.

Labels, pivotal in supervised learning, pair with images to provide strong learning signals. In contrast to images, labels provide highly compressed representations because they are defined in a semantic space. For instance, in the ImageNette-128~\cite{ref_imagenette_imagewang}, representing a ``cassette player'' requires 49,000 scalars ($128\times128\times3$) for the image, but only ten scalars for its one-hot vector label. This substantial difference between image and label suggests a new perspective to dataset distillation, emphasizing the potential of harnessing more information from labels rather than images.

Addressing the overlooked potential of labels in dataset distillation, we introduce Label-Augmented Dataset Distillation (LADD). LADD effectively exploits labels in a distilled dataset. Our approach comprises two main stages: distillation and deployment, as depicted in ~\cref{fig:overview}. In the distillation stage, we first generate synthetic images using existing distillation algorithms. Subsequently, we apply an image sub-sampling algorithm to each synthetic image. For each sub-image (termed a local view), we generate a dense label, sub-image's soft label, which encapsulates high-quality information. During the deployment stage, LADD uniquely merges global view images with their original labels and local view images with the corresponding dense labels, delivering diverse learning signals.

LADD presents three key benefits over prior methods: (1) enhanced storage efficiency with smaller increments in dataset sizes, (2) reduced computational demands, and (3) improved performance and robustness across different testing architectures. First, LADD employs a fixed-parameter sampling rule for sub-image generation, ensuring minimal memory overhead (e.g., only 2.5\% regardless of IPC (images per class)). Second, the computational demands are significantly lowered as the label augmentation process only involves dense label predictions. Lastly, rich information encoded in labels serves as effective and robust training signals at the deployment stage. In this way, LADD leverages the diverse local information obtained from dense labels.

Experimental results validate these key advantages of our LADD. At 5 IPC, LADD consistently surpasses the 6 IPC baseline while consuming 87\% less memory. This underscores the memory efficiency of our method. Additionally, in this setup, LADD only requires an extra 0.002 PFLOPs for label augmentation compared to the 5 IPC baseline. This is notably lower than the additional 211 PFLOPs required by the 6 IPC setup. Furthermore, LADD improves the performances of three baselines by an average of 14.9\%, validated across five test model architectures and five distinct datasets. Finally, GradCAM~\cite{ref_gradcam_selvaraju2017grad} visualizations show that LADD-trained models capture objects within images more accurately. This demonstrates the robustness of our label-augmented distilled dataset approach. 

Our contributions can be summarized as follows:
\begin{itemize}
    \vspace{-0.07in}
    \item We recognize the crucial role of labels in dataset distillation, an aspect neglected in existing research.
    \vspace{-0.1in}
    \item We introduce a novel framework, label-augmented dataset distillation, which utilizes dense labels for local views of each synthetic image. We offer an effective training method for the deployment stage to maximize the use of the distilled dataset.
    \vspace{-0.1in}
    \item Extensive experiments reveal that our method significantly improves computation efficiency, storage efficiency, and cross-architecture robustness. Moreover, our approach can be effectively integrated with existing image-focused distillation methods.
\end{itemize}

\section{Related work}
\label{sec:related}
\noindent\textbf{Preliminary: dataset distillation.}
Dataset distillation is the process of synthesizing a dataset, denoted as $D$, which comprises a small, representative subset of samples extracted from a larger source dataset $D_s$. With the number of total classes $C$ and the number of images per class (\text{IPC}), the distilled dataset $D$ contains $C \times \text{IPC}$ image-label pairs (i.e., $D = \{(x_i, y_i)_{i=1}^{C \times \text{IPC}}\}$).

To achieve dataset distillation, algorithms employ a bi-loop optimization strategy consisting of two phases: the inner-loop and the outer-loop. The inner loop simulates the training of two models with the source dataset $D_s$ and the synthetic dataset $D$, respectively. In detail, two models $f(x_s, \theta_s)$ and $f(x, \theta)$ with the same structure are trained on $D_s$ and $D$ for one or several iterations from the identical initial weights $\theta_O$. Subsequently, with pre-trained models, the outer loop updates the distilled dataset such that the model trained on $D$ approximates the model trained on $D_s$. The optimization objective for the outer loop is to minimize $\mathcal{L}_{sim}$ loss that measures the difference between two trained models at the inner loop:
\begin{equation}
    \mathcal{L}_{sim}(D_s, D) = \text{dist}(f(\cdot;\theta_s), f(\cdot;\theta)).
\end{equation}
Then, the distilled dataset $D$ is updated to reduce the dissimilarity:
\begin{equation}
    D := D - \beta \nabla_{D} \mathcal{L}_{sim}(D_s, D),
\end{equation}
where $\beta$ is the learning rate for the dataset.

We refer to the aforementioned process as the distillation stage. Subsequently, during the deployment stage, we utilize the distilled dataset to train a model, represented as $y = h(x; \phi)$. This model undergoes evaluation on the real validation dataset $D_s^{\text{val}}$.

\noindent\textbf{Trends in dataset distillation algorithm.}
Various distillation methods have been proposed to define the similarity loss, denoted as $L_{sim}$. Performance matching~\cite{ref_dd_wang2018dataset} and distribution matching~\cite{ref_distmat_zhao2023dataset,ref_cafe_wang2022cafe,ref_improveddist_zhao2023improved,ref_datadam_sajedi2023datadam,ref_echo_zhang2023echo} utilize a distance function 
to measure similarity in predictions or features, respectively. 
Gradient matching~\cite{ref_gradmat_zhao2021dataset} aligns gradients of the network parameter $\theta_s$ and $\theta$ for increased efficiency by reducing multiple inner-loop iterations. Trajectory matching~\cite{ref_mtt_cazenavette2022dataset,ref_datm_guo2023towards} focuses on minimizing the parameter distance between $\theta_s$ and $\theta$ after several inner-loop updates. This approach captures the long-range relationship between parameters, an aspect that gradient matching does not address. In contrast, DiM~\cite{ref_dim_wang2023dim} and SRe$^2$L~\cite{ref_sre2l_yin2024squeeze} bypass bi-loop optimization by using conditional GANs and reversing fully-trained models for distilled data synthesis, respectively.

Other methods enhance the robustness or image representation of the distilled dataset. DSA~\cite{ref_dsa_zhao2021dataset} utilizes an augmentation-aware synthesis for diverse image augmentations. ModelAug~\cite{ref_modelaug_zhang2023accelerating} increases the synthesis robustness of $D$ by diversifying the $\theta$ configuration during distillation. AST~\cite{ref_ast_shen2023efficient} uses a smooth teacher in trajectory matching~\cite{ref_mtt_cazenavette2022dataset} to emphasize essential trajectory for $D$ and employs additive noise to augment the teacher while distillation.
To improve image representation, GLaD~\cite{ref_glad_cazenavette2023generalizing} and LatentDD~\cite{ref_latentdd_duan2023dataset} regularize the manifold of $D$ based on GAN~\cite{ref_styleganxl_sauer2022stylegan} and Diffusion Model~\cite{ref_ldm_rombach2022high}. IDC~\cite{ref_idc_kim2022dataset} enriches representation by embedding multiple small images within a single image of $D$.

Our focus is on enriching label space information to enhance distilled dataset quality. We emphasize that our method is both compatible with and capable of synergizing with other distillation methods in image synthesis.

A few methods draw focus to utilizing labels. FDD~\cite{ref_fdd_bohdal2020flexible} optimizes only labels while images are randomly selected from the source dataset. FrePo~\cite{ref_frepo_zhou2022dataset} optimizes both images and labels at once. TESLA~\cite{ref_tesla_cui2023scaling} uses a soft label for each image. These methods are limited to using a single label per image. On the other hand, we augment a single label into multiple informative labels, achieving enhancements in both memory efficiency and performance.




\section{Method}
\label{sec:method}

We propose Label-Augmented Dataset Distillation (LADD), a specialized label augmentation method for dataset distillation. During the dataset distillation stage, LADD conducts a label augmentation process to images distilled by conventional image-level dataset distillation algorithms. For each image $x$, we produce additional groups of soft labels, denoted dense labels, and create a label-augmented dataset $D_{LA}$. Specifically, to obtain $D_{LA}$, the label augmentation step goes through two processes: (1) an image sub-division and (2) a dense label generation. In the deployment stage, LADD uses both global (i.e., full images with hard labels) and local data (i.e., sub-sampled images with dense labels) to train the network effectively. ~\cref{fig:overview} depicts the overview of our method.

In the following section, we describe details of the label augmentation process (\cref{sec:method-LA}) and the labeler acquisition (Sec.~\ref{sec:method-labeler}). Finally, we demonstrate the training procedure of the deployment stage (Sec~\ref{sec:method-deploy}).

\subsection{Label Augmentation}
\label{sec:method-LA}
We denote the image-level distilled dataset $D = \{(x_i, y_i) | i \in [1, C \times \text{IPC}]\}$, where $C$ is the number of classes in the source dataset $D_s$ and $\text{IPC}$ is the number of images per class. In our framework, $D$ is generated using an existing image-level distillation algorithm. By preserving the effectiveness of the image-level distilled dataset, our method synergizes with state-of-the-art dataset distillation algorithms, leveraging their strengths.

\noindent\textbf{Image Sub-Sampling.} We define a function $S$ that samples synthetic image $x_i \in D$ into several sub-images. Considering the memory-constrained environment, dynamic sub-image sampling is not an optimal choice because it requires saving additional sampling parameters. 
Therefore, we restrict $S$ to be a static strategy sampler. We sample $N^2$ sub-images from $x_i$. Each sub-image covers $R\%$ of each axis. To achieve a uniform sampling across $x_i$, we maintain a consistent stride $(100\% - R\%) / (N - 1)$ for cropping. For example, for $x_i$ of $128\times128$ pixels, using $R = 62.5\%$ and $N = 5$, we obtain 25 sub-images of $80\times80$ pixels each, applying a 12-stride. After the sub-sampling, we resize each sub-image to match the dimension of $x_i$. For clarity, we denote the sub-sampling function $S$ as below:
\vspace{-0.05in}
\begin{equation}
    x_{i,j} = S_j(x_i),
\end{equation}
where $j \in [1, N^2]$ is the index of sub-sampled image.

\noindent\textbf{Dense Label Generation.} Sub-images, derived from the same original image, vary in visual content. In detail, each sub-image exhibits distinct patterns, conveying different levels of class information. We generate labels for each sub-image $x_{i,j}$, resulting in $N^2$ labels for each synthetic image $x_i$. To capture rich information in these labels, we opt for soft labeling. We develop the labeler $y^s = g(x)$, where $x$ denotes the image and $y^s$ is the corresponding soft label. We train the labeler on the source dataset $D_s$ from scratch. Then, we obtain a dense label $y^d$ from each sub-image:
\begin{equation}
    y^d_{i,j} = g(S_j(x_i)).
\end{equation}
We will discuss how to train $g$ in Sec~\ref{sec:method-labeler}.

\begin{algorithm}[t]
\caption{Label Augmentation}
\begin{algorithmic}[1]
\State \textbf{Input:} Distilled dataset $D = \{(x_i, y_i)\}$, Labeler $g$, Sub-sampling function $S$
\State \textbf{Output:} Label augmented dataset $D_{LA}$
\For{each image $x_i$ in $D$}
    \For{$j = 1$ to $N^2$}
        \State $x_{i,j} \gets S_j(x_i)$ \Comment{Generate j-th sub-image}
        \State $y_{i,j}^d \gets g(x_{i,j})$ \Comment{Generate sub-image soft label}
    \EndFor
    \State Add $(x_i, y_i, y_{i}^d)$ to $D_{LA}$
\EndFor
\State \textbf{return} $D_{LA}$
\end{algorithmic}
\label{alg:label-augmentation}
\end{algorithm}

After the dense label generation, we obtain the original hard label $y_i$ and a dense label $y^d_i$ containing $N^2$ soft labels for a synthetic image $x_i$. We denote the label augmented dataset as $D_{LA} = \{(x_i, y_i, y^d_i) | i\in[1, C\times\text{IPC}]\}$. The synthesis process of $D_{LA}$ is illustrated in Algorithm~\ref{alg:label-augmentation}.

One straightforward approach might involve optimizing labels as part of the distillation process. However, it adds complexity to an already complicated optimization process, potentially leading to instability. Furthermore, it reduces computational efficiency due to slower convergence and increased operations per iteration. Instead, our LADD first applies existing distillation methods for image-level distillation. Subsequently, we perform a label-augmentation step on the distilled data, producing final datasets with our generated labels. In this way, LADD enjoys significant performance gains with minimal computational overhead.

Both LADD and knowledge distillation~\cite{KD} use a teacher model but differ in the medium of knowledge transfer. Knowledge distillation transfers knowledge through an online teacher during the evaluation stage. However, LADD produces a dataset of images and augmented labels which are fixed after the distillation. In other words, LADD do not require any online model, such as a teacher, during the deployment stage.

\subsection{Acquiring Labeler $g$.}
\label{sec:method-labeler}
LADD employs a labeler $g$ to generate dense labels, employing the same labeler across all evaluations for fairness. To minimize overhead, we design $g$ as a small network mirroring the distillation architecture (ConvNetD5). We train it for 50 epochs with a learning rate of 0.015, saving parameters at epochs 10, 20, 30, 40, and 50.
We use the model trained up to 10 epochs as our early-stage labeler $g$, as it provides general and essential information for sub-images. This is well-aligned with existing dataset distillation methods~\cite{ref_mtt_cazenavette2022dataset,ref_datm_guo2023towards}.
Although $g$ is trained on a source dataset, it appropriately predicts labels for distilled images because the distilled dataset retains local structures of the source data.

Apart from our chosen method, classifiers trained on different data, including zero-shot models like CLIP~\cite{ref_clip_radford2021learning}, can be used as $g$. However, they do not produce more effective dense labels than our method. This is because these pre-trained models are not trained on the distilled dataset and have different architectures from those used in distillation.

\subsection{Training in Deployment Stage}
\label{sec:method-deploy}
We closely follow the deployment stage from existing approaches. Given the dataset $D_{LA}$ and an optimized learning rate $\eta$, we conduct standard classification training on the target network $h(x, \phi)$. Additionally, we modify the data input and training loss to effectively utilize informative dense labels in $D_{LA}$:
\small
\vspace{-0.05in}
\begin{equation}
    L_{cls} = CE(h(x_i, \phi), y_i) + \sum_{j=1}^{N^2}{CE(h(S_j(x_i), \phi), y^d_{i,j})},
\end{equation}
\normalsize
where $CE(\cdot, \cdot)$ is a cross-entropy loss. 
The dimensions of $y_i$ (one-hot) and $y^d_{i,j}$ (soft) are the same as $\mathbb{R}^{C}$, and the dimension of $y^d_{i}$ is $\mathbb{R}^{N^2\times C}$. Through this process, we provide diverse training feedback through augmented dense labels beyond the signal provided by $D$.



\section{Experiment}
\label{sec:exp}

\subsection{Implementation details}
\label{sec:4_1}
\noindent\textbf{Image Sub-Sampling.} The sub-sampling function is selected as a uniform sampler $S$ with $R$ = 62.5\% and $N$ = 5; $R$ and $N$ are determined experimentally (experiments are in Sup.\ref{sup:supA}).
Throughout the experiments, 25 sub-images are generated per synthetic image, and each sub-image is $80\times80$ in size when using $128\times128$ source dataset.
 
\noindent\textbf{Dataset.} Various high-resolution image datasets are used as the source and evaluation datasets. They include ImageNet~\cite{ref_ImageNet_deng2009imagenet} and its subsets, such as ImageNette, ImageWoof~\cite{ref_imagenette_imagewang}, ImageFruit, ImageMeow, and ImageSquawk~\cite{ref_mtt_cazenavette2022dataset}. Each subset contains 10 classes and around 1,300 images per class. All images are center-cropped and resized into $128\times128$.
 
\noindent\textbf{Baselines.}
We benchmark our method against a range of notable dataset distillation methods. These include MTT~\cite{ref_mtt_cazenavette2022dataset}, AST~\cite{ref_ast_shen2023efficient}, GLaD~\cite{ref_glad_cazenavette2023generalizing}, DC~\cite{ref_gradmat_zhao2021dataset}, DM~\cite{ref_distmat_zhao2023dataset}, and TESLA~\cite{ref_tesla_cui2023scaling}. We re-implement DC and DM within the GLaD framework. For all distillation processes, we employ the ConvNetD5, a 5-layer convolutional network~\cite{ref_convdx_gidaris2018dynamic}, as the standard distillation model architecture. For ImageNet-1K, we compare TESLA~\cite{ref_tesla_cui2023scaling}, SRe$^2$L~\cite{ref_sre2l_yin2024squeeze}, and RDED~\cite{ref_rded_sun2023diversity}.

\begin{table}[t]
  
  \centering
  \small
  \resizebox{1.0\columnwidth}{!}{
  \begin{tabular}{c|c|ccc|c} \toprule
IPC                 & Method               & \texttt{MTT}               & \texttt{AST}               & \texttt{GLaD(MTT)}   & Overhead     \\ \midrule
\multirow{3}{*}{1}  & \texttt{Baseline}             & 38.3±0.9          & 39.0±1.2          & 34.3±1.0     & -    \\
                    & \texttt{Baseline++}  & \textbf{42.6±1.0} & \underline{41.8±1.2} & \textbf{41.8±1.4}    & 100.1\%        \\
                    & \texttt{LADD} (ours) & \underline{40.9±1.3} & \textbf{41.9±1.6} & \underline{40.7±1.2} & 2.5\% \\ \midrule
\multirow{3}{*}{5}  & \texttt{Baseline}             & 49.5±1.4          & 51.4±1.2          & 48.0±1.1     & -     \\
                    & \texttt{Baseline++}           & \underline{50.5±1.0}          & \underline{52.1±1.3}          & \underline{48.6±1.2}    & 20.7\%      \\
                    & \texttt{LADD} (ours) & \textbf{52.6±0.8} & \textbf{60.1±0.9} & \textbf{58.4±0.9}  & 2.5\% \\ \midrule
\multirow{3}{*}{10} & \texttt{Baseline}             & 54.6±1.3          & 53.2±0.9          & 52.3±1.1     & -     \\
                    & \texttt{Baseline++}           & \underline{55.4±1.2}          & \underline{54.2±1.3}          & \underline{52.4±1.2}    &  10.0\%      \\
                    & \texttt{LADD} (ours) & \textbf{55.6±1.2} & \textbf{62.0±0.5} & \textbf{62.8±0.9} & 2.5\% \\ \midrule
\multirow{3}{*}{20} & \texttt{Baseline}             & 58.2±1.2          & 55.5±1.5          & 53.3±1.2     & -     \\
                    & \texttt{Baseline++}           & \underline{59.2±1.3}          & \underline{56.9±1.3}          & \underline{54.9±1.0}   & 5.0\%      \\
                    & \texttt{LADD} (ours) & \textbf{59.6±0.5} &
                    \textbf{59.4±1.0}      & \textbf{66.5±0.8}  & 2.5\%   \\ \bottomrule
\end{tabular}%
}
\caption{\textbf{ImageNette (128$\times$128) Performance on Various IPC (images-per-class).} Each result reports an average of validation set accuracy of training ConvNetD5, AlexNet, VGG11, and ResNet18 on synthetic datasets which are distilled using a ConvNetD5 (4-CAE, four cross-architecture evaluation). The numbers after the `±' symbol are the average standard deviation of five trials per evaluation. The best performance is bolded, and the second-best performance is underlined.}
\label{tab:exp-sota-varipc}
\vspace{-0.15in}
\end{table}

\noindent\textbf{Labeler $g$.} To ensure fairness, we use the same labeler $g$ for all experiments. We train $g$ on each source dataset for ten epochs using stochastic gradient descent (SGD) with a learning rate of 0.01 and a batch size of 256, following~\cite{ref_mtt_cazenavette2022dataset}.
 
\noindent\textbf{Cross-Architecture Evaluation.} To evaluate the robustness of distilled data across various architectures, we use five different models~\cite{ref_glad_cazenavette2023generalizing} including four unseen models (ConvNetD5~\cite{ref_mtt_cazenavette2022dataset}, AlexNet~\cite{ref_alexnet_NIPS2012_c399862d}, VGG11~\cite{ref_vgg_DBLP:journals/corr/SimonyanZ14a}, ResNet18~\cite{ref_resnet_he2016residual}, and ViT~\cite{ref_vit_dosovitskiy2020vit}) except in Tab.~\ref{tab:exp-sota-varipc}. We refer to this protocol as 5-CAE. The scores represent the average of five independent trainings for each model. Each test model is trained for 1,000 epochs using the synthetic dataset. We adhere to the learning rate and decay strategy for each model as in~\cite{ref_glad_cazenavette2023generalizing}. Both baseline and LADD use the same data augmentations~\cite{ref_dsa_zhao2021dataset}.

\setlength{\tabcolsep}{4pt}
\begin{table*}[t]
  \centering
  \small
\begin{tabular}{c|cc|cc|cc} \toprule
 & \multicolumn{2}{c|}{\texttt{MTT}} & \multicolumn{2}{c|}{\texttt{AST}} & \multicolumn{2}{c}{\texttt{GLaD(MTT)}} \\
 & \texttt{Baseline}   & \texttt{LADD}       & \texttt{Baseline}   & \texttt{LADD}       & \texttt{Baseline}   & \texttt{LADD}  \\ \midrule
ConvNetD5            & 61.2±1.5   & \textbf{62.1±0.8}   & 63.8±0.5   & \textbf{66.8±0.4}   & 61.2±0.4      & \textbf{69.0±0.8}     \\
VGG11                & 49.6±1.8   & \textbf{50.6±1.5}   & 48.3±1.4   & \textbf{58.1±0.7}   & 49.0±1.0      & \textbf{60.0±1.3}    \\
ResNet18             & 57.3±1.9   & \textbf{59.0±1.6}  & 54.9±0.7   & \textbf{63.6±0.6}   & 55.6±1.9      & \textbf{65.5±0.7}     \\
AlexNet              & 46.4±0.6   & \textbf{51.0±0.6}   & 45.6±1.1   & \textbf{59.4±0.3}   & 43.3±0.9      & \textbf{56.7±0.8}     \\
ViT                  & 35.9±0.8   & \textbf{37.8±0.5}   & 31.0±1.3   & \textbf{32.6±2.2}   & 32.6±0.2      & \textbf{42.5±1.2}     \\ \midrule
Avg.                 & 50.1±1.3   & \textbf{51.8±1.3}   & 48.7±1.0   & \textbf{56.1±0.8}   & 48.3±0.9      & \textbf{58.7±1.0}  \\ \bottomrule  
\end{tabular}%
\caption{\textbf{Detail Results in Cross-Architecture Evaluation.}
   All results are measured on ImageNette dataset at 10 IPC.}
\label{tab:exp-sota-5CAE}
\vspace{-0.1in}
\end{table*}
\begin{figure}[t]
  \centering
  \includegraphics[width=1.\columnwidth]{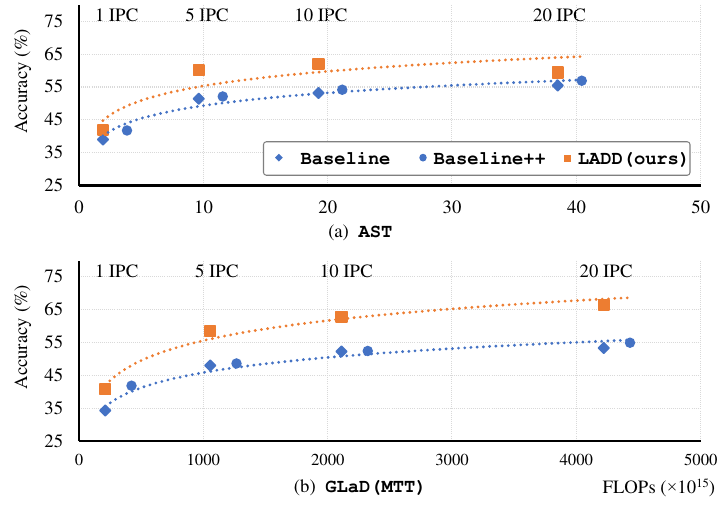}
  \vspace{-0.3in}
  \caption{\textbf{FLOPs-Accuracy Plot for Distillation.} $x$-axis indicates the total computational cost to obtain $D$ in FLOPs. For \texttt{LADD}, we compute FLOPs for both synthesizing $D$ and creating dense labels. Each result uses ImageNette.}
  \label{fig:exp-sota-resource}
  \vspace{-0.2in}
\end{figure}
\begin{figure*}[t]
  \centering
  \resizebox{0.7\textwidth}{!}{\includegraphics{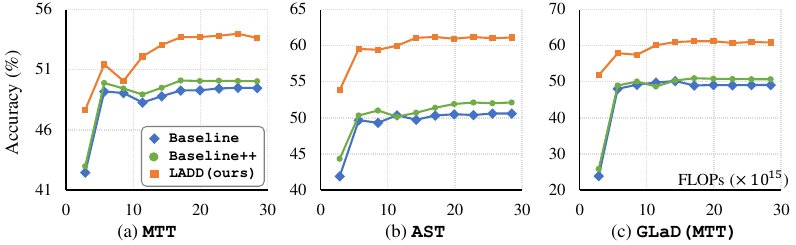}}
  \vspace{-0.07in}
  \caption{\textbf{FLOPs-Accuracy Plot at the Deployment Stage.} $x$-axis indicates the total computational cost at the deployment stage in FLOPs. Among the three algorithms, \texttt{LADD} shows the best performance. Each result uses ImageNette at 5 IPC.}
  \label{fig:exp-fair-train}
  \vspace{-0.1in}
\end{figure*}

\subsection{Quantitative evaluation}
\label{sec:exp-sota}

\begin{table}[t]
\centering
\small
\begin{tabular}{c|cc} \toprule
Method & Accuracy (\%) & Assumption compliance \\ \midrule
\texttt{TESLA}~\cite{ref_sre2l_yin2024squeeze}~\textsubscript{(ICML'23)}     & 7.7±0.1 & {\color{LimeGreen}\ding{52}} \\
\texttt{SRe$ ^2$L}~\cite{ref_sre2l_yin2024squeeze}~\textsubscript{(NeurIPS'23)}     & 21.3±0.6 & {\color{LimeGreen}\ding{52}} \\
\texttt{RDED-I} (H) & 12.4±0.3 & {\color{LimeGreen}\ding{52}} \\
\texttt{RDED-I} (S)   & 23.6±0.3 & {\color{LimeGreen}\ding{52}} \\
\texttt{LADD-RDED-I} (ours) & \textbf{28.8±0.5}   & {\color{LimeGreen}\ding{52}} \\ \midrule
\texttt{RDED}~\cite{ref_rded_sun2023diversity}~\textsubscript{(CVPR'24)}   & 42.0±0.3 &  {\color{BrickRed}\ding{56}} \\ \bottomrule
\end{tabular}%
\caption{\textbf{Performance on ImageNet-1K Dataset.} Each model uses ResNet-18~\cite{ref_resnet_he2016residual} as a test model. IPC is set to 10.}
\vspace{-0.2in}
\label{tab:imnet1k}
\end{table}
\begin{table*}[t]
  \centering
  \small
  \resizebox{1.5\columnwidth}{!}{
  \begin{tabular}{c|ccccc} \toprule
Method                 & ImageNette    & ImageFruit    & ImageWoof     & ImageMeow     & ImageSquawk    \\ \midrule
\texttt{MTT}                    & 45.3±1.1 & 31.7±1.8 & 28.3±1.2 & 33.0±1.1 & 41.5±1.0  \\
\texttt{LADD-MTT} (ours)        & \textbf{49.2±0.9} & \textbf{35.5±1.2} & \textbf{31.0±0.8} & \textbf{36.4±0.7} & \textbf{48.2±0.8}         \\ \midrule
\texttt{AST}                    & 47.3±1.2 & 32.9±1.9 & 29.3±1.1 & 32.0±1.5 & 35.1±2.1 \\
\texttt{LADD-AST} (ours)        & \textbf{53.4±1.1} & \textbf{40.3±1.4} & \textbf{33.0±1.1} & \textbf{36.0±1.0} & \textbf{43.2±1.0}  \\ \midrule
\texttt{GLaD(MTT)}             & 44.2±1.0 & 27.5±1.0 & 24.5±0.9 & 30.0±0.8 & 34.0±1.3 \\
\texttt{LADD-GLaD(MTT)} (ours) & \textbf{53.9±0.9} & \textbf{32.5±1.2} & \textbf{26.1±0.6} & \textbf{33.7±1.1} & \textbf{42.1±0.8} \\ \bottomrule
\end{tabular}%
}
\vspace{-0.07in}
\caption{\textbf{Performance Improvement on Various Datasets.} All methods are trained on each dataset at 5 IPC. All values are 5-CAE results.}
\label{tab:exp-sota-vardata}
\vspace{-0.1in}
\end{table*}


\setlength{\tabcolsep}{7pt}
\begin{table}[t]
  \centering
  \small
\begin{tabular}{c|cc} \toprule
           & \texttt{Baseline} & \texttt{LADD}            \\ \midrule
\texttt{MTT}        & 45.3±1.1 & \textbf{49.2±0.9} \\
\texttt{AST}        & 47.3±1.2 & \textbf{53.4±1.1} \\
\texttt{GLaD(MTT)} & 44.2±1.0 & \textbf{53.9±0.9} \\
\texttt{GLaD(GM)}  & 39.8±0.7 & \textbf{52.1±1.0} \\
\texttt{GLaD(DM)}  & 37.2±1.2 & \textbf{49.9±1.0} \\
\texttt{TESLA}      & 19.2±0.7 & \textbf{27.3±0.7} \\ \bottomrule
\end{tabular}%
\caption{\textbf{Performance on Various Algorithms.}
   All 5-CAE results are measured in ImageNette dataset at 5 IPC.}
\label{tab:exp-sota-baselines}
\vspace{-0.3in}
\end{table}

We quantitatively evaluate \texttt{LADD} by benchmarking it against representative distillation methods (MTT~\cite{ref_mtt_cazenavette2022dataset}, AST~\cite{ref_ast_shen2023efficient}, and GLaD~\cite{ref_glad_cazenavette2023generalizing}) in various IPC settings. \texttt{LADD} incurs additional memory usage compared to the \texttt{baseline} because of labeler training and label augmentation. For fair comparison, we evaluate the baselines with incremented IPC (i.e., IPC+1), labeled as \texttt{baseline++}. We focus on 4-CAE results in ~\cref{tab:exp-sota-varipc} since MTT and AST are not fully compatible with heterogeneous architectures (e.g., several experiments failed to converge on ViT architecture). The additional memory overhead for both images (\texttt{uint8}) and labels (\texttt{float32}) is calculated utilizing the Python \textit{zipfile} library~\cite{ref_reb_zip}, the standard compression method.

Tab.~\ref{tab:exp-sota-varipc} presents the results for varying IPC on the ImageNette. The quantitative analysis reveals that \texttt{LADD} surpasses the \texttt{baseline}, showing an average improvement of 15\% at 5 IPC. Notably, our method outperforms \texttt{baseline++} in all cases except at 1 IPC. 
At 1 IPC, \texttt{baseline++} entails a 100.1\% increase in memory usage. In contrast, \texttt{LADD} achieves comparable performance with only a 2.5\% storage overhead, resulting in 40 times greater memory efficiency. For 5 IPC, \texttt{baseline++} requires 20.7\% more memory to accommodate an extra image per class. Conversely, \texttt{LADD} requires only an additional 2.5\% memory while achieving, on average, a 13.2\% better performance than \texttt{baseline++} across three models. Consequently, we conclude that our approach shows impressive performances in terms of accuracy and efficiency, creating synergies with existing dataset distillation algorithms.

We evaluate the cross-architecture robustness of our method. Tab.~\ref{tab:exp-sota-5CAE} shows results for five architectures during the deployment stage. Notably, the baseline's ViT exhibits the weakest performance due to the architectural divergence between the models in the distillation and deployment stages. Therefore, ViT's performance is a key indicator of the architecture robustness of the distilled dataset. \texttt{LADD} enhances performance across various architectures, particularly boosting ViT performance by 31\% in \texttt{GLaD(MTT)}. The dense label in \texttt{LADD} improves the representation quality and generalization within the distilled dataset.

Additionally, we show that LADD surpasses other dataset distillation methods on the ImageNet-1K~\cite{ref_ImageNet_deng2009imagenet}, as shown in Tab.~\ref{tab:imnet1k}. ImageNet-1K presents significant challenges in dataset distillation due to high GPU consumption and complex optimization.
For RDED, we remove the labeling process that uses the teacher model at the deployment stage. Using the teacher model at deployment stage violates the assumption of dataset distillation because it aligns more with knowledge distillation (\cref{sec:method-LA}). We denote the modified model as \texttt{RDED-I} (H or S), which consists of the distilled image and either hard or soft labels. Without online knowledge transfer of the \texttt{RDED}, we observe that \texttt{RDED-I} (H) only achieves 12.4\% accuracy. \texttt{RDED-I} (S) shows better accuracy at 23.6\%, which is better than \texttt{SRe$ ^2$L}. Our method demonstrates the best performance. We conclude that our approach improves the performance on a large dataset. More details are described in the Sup.\ref{sup:supB}.

We compute the FLOPs requirement to assess the computational overhead for creating distilled data $D$ and $D_{LA}$. Fig.~\ref{fig:exp-sota-resource} presents the total FLOPs necessary to distill $D$ ($\blacklozenge$, \scalebox{1.4}{$\bullet$}) and $D_{LA}$ ($\blacksquare$). It also shows their corresponding deployment stage accuracies for \texttt{baseline}, \texttt{baseline++}, and \texttt{LADD}. Our observations indicate that \texttt{LADD} is more resource-efficient and achieves higher accuracy than both \texttt{baseline} and \texttt{baseline++}. There's a noticeable offset between the trend lines of \texttt{LADD} and \texttt{baseline}.
This difference highlights our greater computational efficiency compared to previous studies.
According to Fig.~\ref{fig:exp-sota-resource}, the computational cost of LADD is slightly higher than that of the \texttt{baseline}, but significantly lower than that of \texttt{baseline++}. This is because LADD's computation includes labeler training and label augmentation in addition to the baseline distillation. However, these additional costs are much smaller than those for baseline distillation. Thus, it is a fair comparison of computational efficiency.

Furthermore, for an equitable comparison of the training cost, we conduct the experiments using the same batch size and number of iterations during the deployment stage. Fig.~\ref{fig:exp-fair-train} depicts the accuracy of each model relative to the training cost. LADD outperforms both the \texttt{baseline} and \texttt{baseline++} under the same training cost.

In Tab.~\ref{tab:exp-sota-vardata}, we report performances across various datasets. These results consistently demonstrate that \texttt{LADD} significantly enhances the performance of baselines across different source datasets. 
For each baseline model, we calculated the percentage improvement of \texttt{LADD} over the original models for all five datasets and then averaged them. We further averaged the improvements across the three baselines. This comprehensive calculation shows that \texttt{LADD} achieves an average performance improvement of 14.9\% across the five datasets. This consistent improvement is a strong indication of our method's generalizability, regardless of the dataset. Tab.~\ref{tab:exp-sota-baselines} presents the results from using various distillation algorithms. Analogous to the previous results, \texttt{LADD} significantly outperforms the various \texttt{baselines}. \texttt{TESLA} depicts low accuracy in both Tab.~\ref{tab:imnet1k} and Tab.~\ref{tab:exp-sota-baselines} because it reduces computations by ignoring training feedback. Detailed information is described in the Sup.\ref{sup:supC}.
Based on the experiments, we conclude that \texttt{LADD} demonstrates robustness and efficiency across a range of IPC settings, datasets, and architectures.

In conclusion, our extensive experiments establish that our method is effective in several key aspects. First, it demonstrates resource efficiency, as illustrated in Fig.~\ref{fig:exp-sota-resource}. Second, it provides high compactness relative to its performance, evidenced in Tab.~\ref{tab:exp-sota-varipc}. Third, it consistently delivers superior training performance in diverse environments, as shown in Tab.~\ref{tab:exp-sota-vardata} and \ref{tab:exp-sota-baselines}. These findings collectively confirm that \texttt{LADD} significantly improves the quality of distilled datasets via efficient label augmentation.

\subsection{Impact of Dense Labels in LADD}
\label{sec:exp-aug-eval}

\begin{table*}[t]
\centering
\small
\resizebox{0.9\textwidth}{!}{
\begin{tabularx}{\textwidth}{@{\extracolsep{\fill}} cc|cc|ccccc|c}
\toprule
\multicolumn{2}{c|}{Images} & \multicolumn{2}{c|}{Labels} & \multirow{2}{*}{ConvNetD5} & \multirow{2}{*}{VGG11} & \multirow{2}{*}{ResNet18} & \multirow{2}{*}{AlexNet} & \multirow{2}{*}{ViT} & \multirow{2}{*}{Avg.} \\
Full & Sub-sampled & Hard & Soft & & & & & & \\
\midrule
\checkmark & & \checkmark & & 58.7 & 45.5 & 50.6 & 37.0 & 29.4 & 44.2 \\
\checkmark & & & \checkmark & 60.1 & 44.5 & 51.2 & 37.7 & 28.8 & 44.5 \\
\checkmark & & \checkmark & \checkmark & 60.8 & 44.1 & 51.9 & 36.3 & 29.2 & 44.5 \\
& \checkmark & \checkmark & & 54.3 & 49.7 & 49.5 & 37.3 & 29.6 & 44.1 \\
& \checkmark & & \checkmark & 62.5 & 53.8 & 57.0 & 49.4 & 32.6 & 51.1 \\
& \checkmark & \checkmark & \checkmark & 59.8 & 54.7 & 55.4 & 48.9 & 34.6 & 50.7 \\
\checkmark & \checkmark & \checkmark & \checkmark & \textbf{66.5} & \textbf{55.7} & \textbf{61.2} & \textbf{50.2} & \textbf{35.9} & \textbf{53.9} \\
\bottomrule
\end{tabularx}
}
\caption{\textbf{Performance Analysis on Image and Label Combinations.} \texttt{GLaD(MTT)} is set to the baseline model. All results are 5-CAE values measured on ImageNette at 5 IPC.}
\vspace{-0.2in}
\label{tab:exp-aug-eval}
\end{table*}

In this section, we investigate the most efficient ways to utilize a distilled dataset. We designate \texttt{GLaD(MTT)} as our baseline model. Tab.~\ref{tab:exp-aug-eval} presents the deployment stage performance using different combinations of datasets and labels. We note that the performance differences are negligible when training each image in $D$ with hard labels, soft labels, or a mix of both. Additionally, using only sub-images with hard labels yields results comparable to the baseline. However, employing sub-images with corresponding dense labels results in a significant performance improvement of 7\%p. This underscores that the combined strategy of image sub-sampling and dense label generation in \texttt{LADD} is highly effective for label utilization. Furthermore, integrating training with full images and their hard labels into previous experiments leads to an extra 2.8\%p boost. This demonstrates that \texttt{LADD}, which leverages both local views with dense labels and global views of distilled images, is the most effective approach for label augmentation.

\subsection{Dataset Quality Analysis}
\label{sec:exp-quali-anal}

\begin{figure}[t]
  \centering
  \includegraphics[width=1\columnwidth]{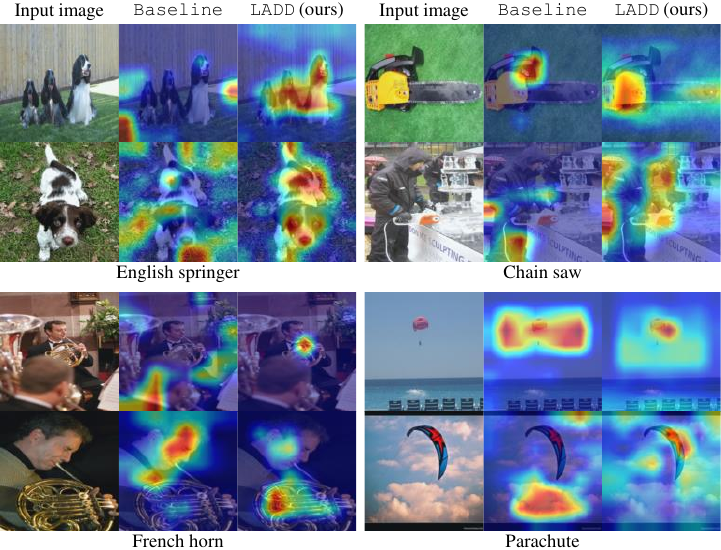}
  \caption{\textbf{Analysis on the Dataset Quality.} The second and third columns depict GradCAM~\cite{ref_gradcam_selvaraju2017grad} visualization of each prediction from \texttt{GLaD(MTT)} (\texttt{baseline}) and \texttt{LADD-GLaD(MTT)} (\texttt{LADD}), respectively.}
  \label{fig:dqa-gradcam}
  \vspace{-0.23in}
\end{figure}

We employ GradCAM~\cite{ref_gradcam_selvaraju2017grad} to visually investigate the reasons behind performance improvements from label augmentation. Fig.~\ref{fig:dqa-gradcam} displays the GradCAM results for \texttt{GLaD(MTT)} and \texttt{LADD}, both trained on ImageNette at 5 IPC. Our observations reveal that \texttt{LADD} more accurately identifies objects than the \texttt{baseline}, which often focuses on surroundings rather than primary objects. For example, \texttt{LADD} effectively concentrates on the main object, identifying all three English springers. Another shortcoming of the \texttt{baseline} is its tendency to detect only parts of an object, while \texttt{LADD} captures entire objects for accurate classification. Additionally, \texttt{LADD} excels at detecting small objects like a miniature French horn and a Parachute, outperforming the \texttt{baseline}. Overall, models trained with \texttt{LADD} classify objects with diverse features better, regardless of size, quantity, and structure. This demonstrates \texttt{LADD}'s ability to learn multiple representations of a single object using diverse dense labels with sub-images, significantly enhancing classification accuracy. Challenging categories like Chain saw, French horn, Gas pump, and Golf ball are difficult to classify (accuracies $\leq$ 40\%)
due to variations in size and quantity. \texttt{LADD} improves classification accuracies from 32\%, 36\%, 32\%, and 40\% to 56\%, 60\%, 40\%, and 56\%, respectively, marking up to a 24\% improvement.

\subsection{Ablation Study}
\label{sec:exp-labeler}
\begin{figure}[t]
  \centering
  \includegraphics[width=1.\columnwidth]{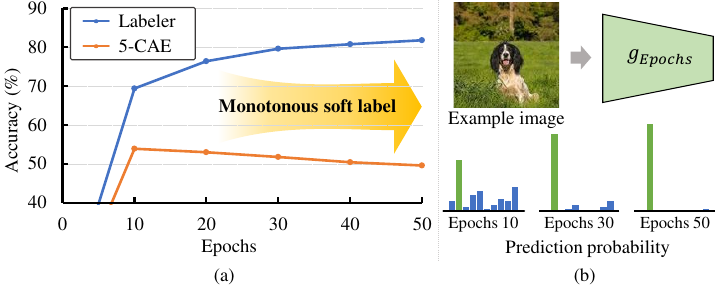}
  \vspace{-0.2in}
  \caption{\textbf{Analysis on the Labeler $g$.} (a) The Blue line indicates the labeler performance. The orange line depicts the accuracy of the test model in the deployment stage where dense labels in the distilled dataset are obtained from the labeler of each epoch. (b) Each bar graph depicts the prediction probability of the example image using the labeler for each epoch.}
  \label{fig:exp-labeler-logit}
  \vspace{-0.1in}
\end{figure}

The ablation study on \texttt{LADD-GLaD(MTT)} using the ImageNette at 5 IPC concentrates on identifying the ideal training steps for the labeler.
The labeler creates soft labels that encapsulate meaningful information for specific sub-images. We evaluate the contribution of training labeler on the source dataset to the distilled dataset. Fig.~\ref{fig:exp-labeler-logit} (a) displays the performance of labeler and \texttt{LADD} across various training epochs. Fig.~\ref{fig:exp-labeler-logit} (b) shows that soft labels from less extensively trained labelers exhibit greater diversity (indicating less overconfidence) compared to those trained for longer periods. This occurs as, during initial training stages, the model primarily absorbs general information about the source dataset. Subsequently, the model begins to memorize the training data, leading to overconfident results. Consequently, we employ a labeler trained only for ten epochs, capitalizing on this early-stage learning.

\section{Conclusion and Limitation}

In this work, we highlight the overlooked role of labels in distilled datasets. Addressing this limitation, we introduce Label-Augmented Dataset Distillation (LADD), a method that effectively utilizes labels. Our approach enriches labels with useful information, orthogonal to the images. This yields three major advantages: (1) enhanced efficiency in distillation computation, (2) improved memory capacity efficiency, and (3) increased dataset robustness.

Extensive experiments demonstrate that LADD enhances various distillation methods with minimal extra computational and memory resources. On five ImageNet subsets and three baseline methods, LADD achieves an average performance improvement of 14.9\% with only a 2.5\% memory increase. Remarkably, LADD surpasses baselines with more images per class while using fewer computational resources and memory capacity. LADD with 5 IPC delivers 12.9\% more accuracy than a 6 IPC baseline while using eight times less memory. We confirmed that datasets distilled using LADD enable more robust training across diverse architectures. Additionally, results from GradCAM~\cite{ref_gradcam_selvaraju2017grad} visualizations show that models trained with our dataset accurately and robustly capture object locations.

\noindent\textbf{Limitation.} Our approach requires training a labeler to generate dense labels, which may need extra resources. However, this is more efficient than re-distilling the dataset with more images per class. Once trained, the labeler continuously produces dense labels for the same dataset.





{\small
\bibliographystyle{ieee_fullname}
\bibliography{egbib}
}

\appendix

\twocolumn[
    \begin{center}
        \LARGE{\bf{Supplementary Material for Label-Augmented Dataset Distillation}}
    \end{center}
]

\setcounter{table}{0}
\setcounter{figure}{0}
\setcounter{equation}{0}
\renewcommand{\thetable}{S\arabic{table}}
\renewcommand{\thefigure}{S\arabic{figure}}
\renewcommand{\theequation}{S\arabic{equation}}

\section{Sub-Sampling Hyperparameter $N$ and $R$}
\label{sup:supA}
\setlength{\tabcolsep}{5pt}
\begin{table*}[t]
    \centering
        \begin{tabular}{c|cccc|c}
            \toprule
                 \backslashbox{R (pixels)}{N} & 3    & \underline{5}    & 7    & 9    & Avg. \\ \midrule
                50.0\% (64) & 47.0±1.0 & 53.4±0.9 & 55.0±0.7 & 56.3±0.8 & 52.9±0.9 \\
                 \underline{62.5\% (80)} & 48.8±1.3 & 53.9±0.9 & 55.2±1.3 & 54.2±1.0 & \textbf{53.0±1.1} \\
                75.0\% (96) & 48.9±0.9 & 52.1±1.5 & 51.4±1.5 & 52.0±1.2 & 51.1±1.3 \\
                88.5\% (128) & 48.4±1.8 & 50.5±1.2 & 50.9±1.1 & 51.6±1.0 & 50.4±1.3 \\ \midrule
                Avg.   & 48.3±1.3 & 52.5±1.1 & 53.1±1.2 & \textbf{53.5±1.0} &     \\ \midrule
                Overhead (\%)   & 7.5 & 20.7 & 40.2 & 66.3 &     \\\bottomrule
        \end{tabular}
        \caption{\textbf{Ablation Study on Sub-Image Size $R(\%)$ and the Number of Axis Split $N$}. Each accuracy indicates \texttt{LADD-GLaD(MTT)} results on ImageNette at 5 IPC. Underline depicts chosen parameter for other experiments.}
    \vspace{-0.12in}
        \label{tab:exp-subimg-RN}
\end{table*}

We perform the study on \texttt{LADD-GLaD(MTT)} using the ImageNette dataset at 5 IPC. It aims to determine the optimal size ($R$) and the number ($N$) of image sub-samplings. We test four different sizes $R$ and quantities $N$, validating \texttt{LADD-GLaD(MTT)} with 5-CAE. Tab.~\ref{tab:exp-subimg-RN} shows that an increase in $N$ correlates with improved overall accuracy. This is expected, as a higher number of soft labels in a dense label encompasses more information. However, increasing $N$ also results in greater memory inefficiency. For instance, comparing $N=5$ with $N=7$, the performance gain is a mere 1.1\%, but the overhead rises by 94\%. Therefore, balancing the performance-efficiency trade-off is crucial. Hence, we select $N$ = 5 for our model, considering both performance and efficiency.

$R$ represents the size of the sub-image. If $R$ is too small, vital objects representing the target class may be absent in most sub-images. This results in performance degradation due to information loss. Conversely, if $R$ is too large, label augmentation efficiency drops because of redundant information in each sub-image. Our observations indicate that $R$ = 62.5\% yields the most accurate results. Therefore, we choose $R$ = 62.5\% for our model.

\section{Fair Comparison Settings for RDED}
\label{sup:supB}
RDED~\cite{ref_rded_sun2023diversity} introduces an efficient approach for distilling large-scale datasets. It achieves a remarkable 42\% top-1 validation accuracy with ResNet-18~\cite{ref_resnet_he2016residual} on the ImageNet-1K dataset~\cite{ref_ImageNet_deng2009imagenet}. RDED first generates diverse and realistic data through an optimization-free algorithm backed by $\nu$-information theory~\cite{ref_nuinfo_Xu2020A}, which is equivalent to the distillation step. In the deployment stage, the method augments the distilled images and computes the corresponding soft labels from the teacher model. Then, it trains the test model using the augmented images and soft labels.

Despite the remarkable performance of RDED, we identified that the method does not align with the purpose of dataset distillation. Dataset distillation aims to distill the knowledge from a given dataset into a terse data summary~\cite{ref_ddsurvey_sachdeva2023data}. However, RDED uses a teacher model for soft label prediction of augmented images in the deployment stage. Specifically, RDED generates an unlimited number of images and labels via image augmentation that fully exploits the teacher model's knowledge. Thus, RDED aligns more with knowledge distillation rather than dataset distillation in the deployment stage.


Therefore, we assess the performance of RDED while ensuring it complies with the purpose of dataset distillation by eliminating the labeling process that relies on the teacher model during the deployment stage. 

\section{Performance Degradation in TESLA}
\label{sup:supC}
TESLA consistently depicts low accuracy in both Tab.~\ref{tab:imnet1k} and Tab.~\ref{tab:exp-sota-baselines}. Although we used the official code and tuned the hyperparameters, we could not successfully train TESLA. Thus, we investigated the reason for this result.

TESLA introduces a method to reduce the high GPU memory issue arising from the bi-loop nested optimization problem in MTT~\cite{ref_mtt_cazenavette2022dataset}. Through a formulaic improvement, it reduces unnecessary computation graphs while achieving the same objective. Specifically, TESLA claims that the gradient for each batch image only depends on the iteration involving the images. Thereby, the model can remove the computation graph after computing the gradient for each image.

We found an oversight in TESLA’s formulation: it does not consider the inner-loop model parameters as dependent variables of the image from different iterations. This means TESLA simplifies the objective of MTT by ignoring the feedback from different training iterations to reduce computations. This explains why TESLA is incapable of achieving a similar high accuracy to MTT in Tab.~\ref{tab:imnet1k}. Detailed proof can be found in Sec.~\ref{sec:suppl-tesla}.

\section{Experiments on Small Dataset: CIFAR-10}

\setlength{\tabcolsep}{7pt}
\begin{table}[h]
\small
\centering
\begin{tabular}{c|cc} \toprule
IPC       & 1        & 10          \\ \midrule
Random    & 15.4±0.3 & 31.0±0.5   \\
\texttt{DC}~\cite{ref_gradmat_zhao2021dataset}   & 28.3±0.5 & 44.9±0.5   \\
\texttt{DM}~\cite{ref_distmat_zhao2023dataset}   & 26.0±0.8 & 48.9±0.6    \\
\texttt{DSA}~\cite{ref_dsa_zhao2021dataset}      & 28.8±0.7 & 52.1±0.5    \\
\texttt{CAFE}~\cite{ref_cafe_wang2022cafe}      & 30.3±1.1 & 46.3±0.6     \\
\texttt{FRePo}~\cite{ref_frepo_zhou2022dataset}  & 46.8±0.7 & 65.5±0.4  \\
\texttt{MTT}~\cite{ref_mtt_cazenavette2022dataset}  & 46.2±0.8 & 65.4±0.7   \\
\texttt{FTD}~\cite{ref_FTD_du2023minimizing}       & 46.0±0.4 & 65.3±0.4     \\
\texttt{DATM}~\cite{ref_datm_guo2023towards}      & 46.9±0.5 & 66.8±0.2    \\ \midrule
\texttt{DATM}$^\dag$  & 47.6±0.3   & 65.5±0.5      \\
\texttt{LADD-DATM}  (ours) & \textbf{48.6±0.7} & \textbf{67.2±0.4}   \\ \bottomrule
\end{tabular}
\caption{\textbf{Performance on CIFAR-10 Dataset.} \texttt{DATM}$^\dag$ indicates the performance of the reproduced image which is used in \texttt{LADD-DATM}. 
}
\label{tab:suppl-cifar10}
\end{table}

We evaluate LADD on the small-sized image dataset, CIFAR-10~\cite{ref_CIFAR_krizhevsky2009learning}. We adopt the same hyperparameters (i.e., R and N) defined in Sec.~\ref{sec:4_1}, with an image size of $32\times32$. We apply LADD to the distilled dataset from DATM~\cite{ref_datm_guo2023towards}, which is the current state-of-the-art method for small-sized datasets. To account for the small-sized image, we use a 3-layer convolutional network (ConvNetD3) for both the distillation and deployment stages. Tab.~\ref{tab:suppl-cifar10} reports the deployment stage performance at 1 and 10 IPC. The results demonstrate that our method improves \texttt{DATM} and achieves the highest performance compared to other methods. Therefore, we conclude that LADD also boosts performance in small-sized datasets.

\section{Mathematical Analysis on TESLA}
\label{sec:suppl-tesla}

In this section, we derive the mathematical differences between TESLA and MTT to explain the performance difference in Tab.~\ref{tab:imnet1k} and Tab.~\ref{tab:exp-sota-baselines}. 

\subsection{Objective Function of MTT}

We briefly review the mathematical expression of MTT to understand the oversight in TESLA.
MTT defines the $\mathcal{L}_{sim}$ through the parameter distance:
\begin{equation}
    \label{eq:tm_loss}
    \mathcal{L}_{sim} = \| \hat{\theta}_{t+T} - \theta_{t+M}^* \|^2_2 / \| \theta_t^{*} - \theta_{t+M}^* \|_2^2,
\end{equation}
where $\theta_t^*$ and $\theta_{t+M}^*$ are the model parameters trained on source dataset $D_s$ for $t$ and $t+M$ steps, respectively. Starting from the $\theta_t^*$, MTT trains the model for $i \in [0, T)$ steps on the distilled dataset $D$ following the SGD rule and cross-entropy loss. The trained parameter is denoted as:
\begin{equation}
\label{eq:tm_update}
\hat{\theta}_{t+i+1} = \hat{\theta}_{t+i} - \beta\nabla_\theta\ell(\hat{\theta}_{t+i}; \tilde{X}_{i}), 
\end{equation}
where $\tilde{X}_{i}$ is sub-batch of $D$ and $\ell(\hat{\theta}_{t+i}; \tilde{X}_{i})$ is the cross-entropy loss. $\beta$ indicates the learning rate for the inner-loop. We can expand $\hat{\theta}_{t+T}$ as:
\begin{align}
    \hat{\theta}_{t+T} &= \theta_{t}^* - \beta\nabla_{\theta} \ell(\theta_{t}^{*};\tilde{X}_0) - \beta\nabla_{\theta} \ell(\hat{\theta}_{t+1};\tilde{X}_{1}) - ... \nonumber\\
    &- \beta\nabla_{\theta} \ell(\hat{\theta}_{t+T-1};\tilde{X}_{T-1}). \label{eq:expansion}
\end{align}

Eqn.~\ref{eq:tm_loss} is expanded as:
\begin{align}
    \label{eq:square_loss_ref}
    \|\hat{\theta}_{t+T}-\theta_{t+M}^*\|^2_2= \nonumber \\ &
    \|\theta_{t}^* - \beta\overset{T-1}{\underset{i=0}{\sum}}\nabla_{\theta} \ell(\hat{\theta}_{t+i};\tilde{X}_i) - \theta_{t+M}^*\|^2_2.
\end{align}
We omit the constant denominator of $\mathcal{L}_{sim}$ for brevity. We then further expand the Eqn.~\ref{eq:square_loss_ref} as:
\begin{align}
\|\hat{\theta}_{t+T} - \theta_{t+M}^*\|_2^2 \ = \ & 
 2\beta(  \theta_{t+M}^* - \theta_t^*)^T(\overset{T-1}{\underset{i=0}{\sum}}\nabla_{\theta} \ell(\hat{\theta}_{t+i};\tilde{X}_i))
  \nonumber \\  +\beta^2 &\|\sum_{i=0}^{T-1} \nabla_\theta \ell(\hat{\theta}_{t+i}; \tilde{X}_i) \|^2 + C, \label{eq:111}
\end{align}
where $C=\|\theta_{t}^* - \theta_{t+M}^*\|_2^2$ is a constant and a negligible term in the gradient computation. For convenience, we represent $G=\sum_{i=0}^{T-1} \nabla_{\theta} \ell(\hat{\theta}_{t+i};\tilde{X}_{i})$.

\subsection{Cause of Performance Degradation}

TESLA claims two points. First, the elements of the first term $G$ only involve the gradients in a single batch and thus can be pre-computed. Second, the computation graph of $ \nabla_{\theta} \ell(\hat{\theta}_{t+i};\tilde{X}_{i}) $ is not required in the derivative of any other batch $\tilde{X}_{j \neq i}$. Based on these points, TESLA computes the gradient for each batch $\tilde{X}_{i}$ as:
\begin{align}
\frac{\partial \|\hat{\theta}_{t+T}-\theta^*_{t+M}\|_2^2}{\partial \tilde{X}_i} = & \ 2\beta(\theta_{t+M}^* - \theta_t^*)^T \frac{\partial}{\partial \tilde{X}_i}\nabla_{\theta} \ell(\hat{\theta}_{t+i};\tilde{X}_i) \nonumber\\
&+ 2\beta^2 G^T \frac{\partial}{\partial \tilde{X}_i} \nabla_{\theta} \ell(\hat{\theta}_{t+i};\tilde{X}_{i}). \label{eq:gradient} 
\end{align}
Since Eqn.~\ref{eq:gradient} can be computed for each batch, TESLA asserts that the memory requirement can be significantly reduced by not retaining the computation graph for all batches.

Here, we found the missing point in the second claim. The computation graph of $\nabla_{\theta} \ell(\hat{\theta}_{t+i};\tilde{X}_{i})$ is required in the derivative of any other batch $\tilde{X}_{j \neq i}$. For example, we can compute the gradient for $\tilde{X}_{T-2}$ from the Eqn.~\ref{eq:111}:

\begin{align}
& \frac{\partial \|\hat{\theta}_{t+T}-\theta^*_{t+M}\|_2^2}{\partial \tilde{X}_{T-2}} = \nonumber \\
& 2\beta(\theta_{t+M}^* - \theta_t^*)^T \frac{\partial}{\partial \tilde{X}_{T-2}}
\left[\nabla_{\theta} \ell(\hat{\theta}_{t+T-1};\tilde{X}_{T-1}) \right. \nonumber \\
& \ \ \ \ \ \ \ \ \ \ \ \ \ \ \ \ \ \ \ \ \ \ \ \ \ \ \ \ \ \ \ \ \ \ \ \ \ \ \ \ \ \ \left. + \nabla_{\theta} \ell(\hat{\theta}_{t+T-2};\tilde{X}_{T-2})\right] \nonumber \\
& + 2\beta^2 G^T \frac{\partial}{\partial \tilde{X}_{T-2}} \left[\nabla_{\theta} \ell(\hat{\theta}_{t+T-1};\tilde{X}_{T-1}) \right. \nonumber \\
& \ \ \ \ \ \ \ \ \ \ \ \ \ \ \ \ \ \ \ \ \ \ \ \ \ \ \ \ \ \ \ \ \ \left.  + \nabla_{\theta} \ell(\hat{\theta}_{t+T-2};\tilde{X}_{T-2})\right].
\label{eq:gradient_correct}
\end{align}

We can omit other $\nabla_{\theta} \ell(\hat{\theta}_{t+i};\tilde{X}_{i})$ where $i < T - 2$ because they are independent of $\tilde{X}_{T-2}$. However, the term $\nabla_{\theta} \ell(\hat{\theta}_{t+T-1};\tilde{X}_{T-1})$ cannot be ignored. Following Eqn.~\ref{eq:tm_update}, $\hat{\theta}_{t+T-1}$ depends on the synthetic image $\tilde{X}_{T-2}$. The derivative for $\hat{\theta}_{t+T-1}$ with respect to the image is:
\begin{align}
\frac{\partial}{\partial \tilde{X}_{T-2}} \hat{\theta}_{t+T-1} = - \beta \frac{\partial}{\partial \tilde{X}_{T-2}} \nabla_{\theta} \ell(\hat{\theta}_{t+T-2};\tilde{X}_{T-2}).
\end{align}
Then, we can compute the derivative for the term $\nabla_{\theta} \ell(\hat{\theta}_{t+T-1};\tilde{X}_{T-1})$:
\footnotesize
\begin{align}
 \frac{\partial}{\partial \tilde{X}_{T-2}} \nabla_{\theta} & \ell(\hat{\theta}_{t+T-1};\tilde{X}_{T-1}) \nonumber \\
& = \nabla^2_\theta \ell(\hat{\theta}_{t+T-1};\tilde{X}_{T-1}) \frac{\partial}{\partial \tilde{X}_{T-2}} \hat{\theta}_{t+T-1} \nonumber \\
& = - \beta \nabla^2_\theta \ell(\hat{\theta}_{t+T-1};\tilde{X}_{T-1}) \frac{\partial}{\partial \tilde{X}_{T-2}} \nabla_{\theta} \ell(\hat{\theta}_{t+T-2};\tilde{X}_{T-2}).
\end{align}
\normalsize
Finally, the Eqn.~\ref{eq:gradient_correct} becomes:
\small
\begin{align}
& \frac{\partial \|\hat{\theta}_{t+T}-\theta^*_{t+M}\|_2^2}{\partial \tilde{X}_{T-2}} =&  \nonumber \\
& A\left(1 - \beta \nabla^2_\theta \ell(\hat{\theta}_{t+T-1};\tilde{X}_{T-1}) \right) \frac{\partial}{\partial \tilde{X}_{T-2}} \nabla_{\theta} \ell(\hat{\theta}_{t+T-2};\tilde{X}_{T-2}),
\end{align}
\normalsize
where $A = 2\beta(\theta_{t+M}^* - \theta_t^*)^T + 2\beta^2 G^T $. It is obvious that the computation graph of $\nabla_{\theta} \ell(\hat{\theta}_{t+T-1};\tilde{X}_{T-1})$ is required to compute the gradient for $\tilde{X}_{T-2}$. In general, the correct gradient for each batch $\tilde{X}_{i}$ is:
\begin{flalign}
& \frac{\partial \|\hat{\theta}_{t+T}-\theta^*_{t+M}\|_2^2}{\partial \tilde{X}_{i}} = \nonumber \\
& A \prod_{j=i}^{T-1} \left(1 - \beta \nabla^2_\theta \ell(\hat{\theta}_{t+j};\tilde{X}_{j}) \right) \frac{\partial}{\partial \tilde{X}_{i}} \nabla_{\theta} \ell(\hat{\theta}_{t+i};\tilde{X}_{i}).
\label{eq:final}
\end{flalign}
Due to the product term in Eqn.~\ref{eq:final}, the computation graphs for other steps are required to compute the gradient of $\tilde{X}_{i}$. 

In conclusion, the assumption in Eqn.~\ref{eq:gradient} of TESLA neglects that the $\tilde{X}_{i}$ affects the other batch gradients. We also empirically confirm that the gradients for distilled images computed on MTT and TESLA are not identical when all other parameters (such as input distilled images, starting parameters, and learning rates) are equal. We conjecture that the low performance of TESLA is due to this observation.

\section{Visualization of Sub-Samples}
\begin{figure}[h]
  \centering
  \includegraphics[width=1.\columnwidth]{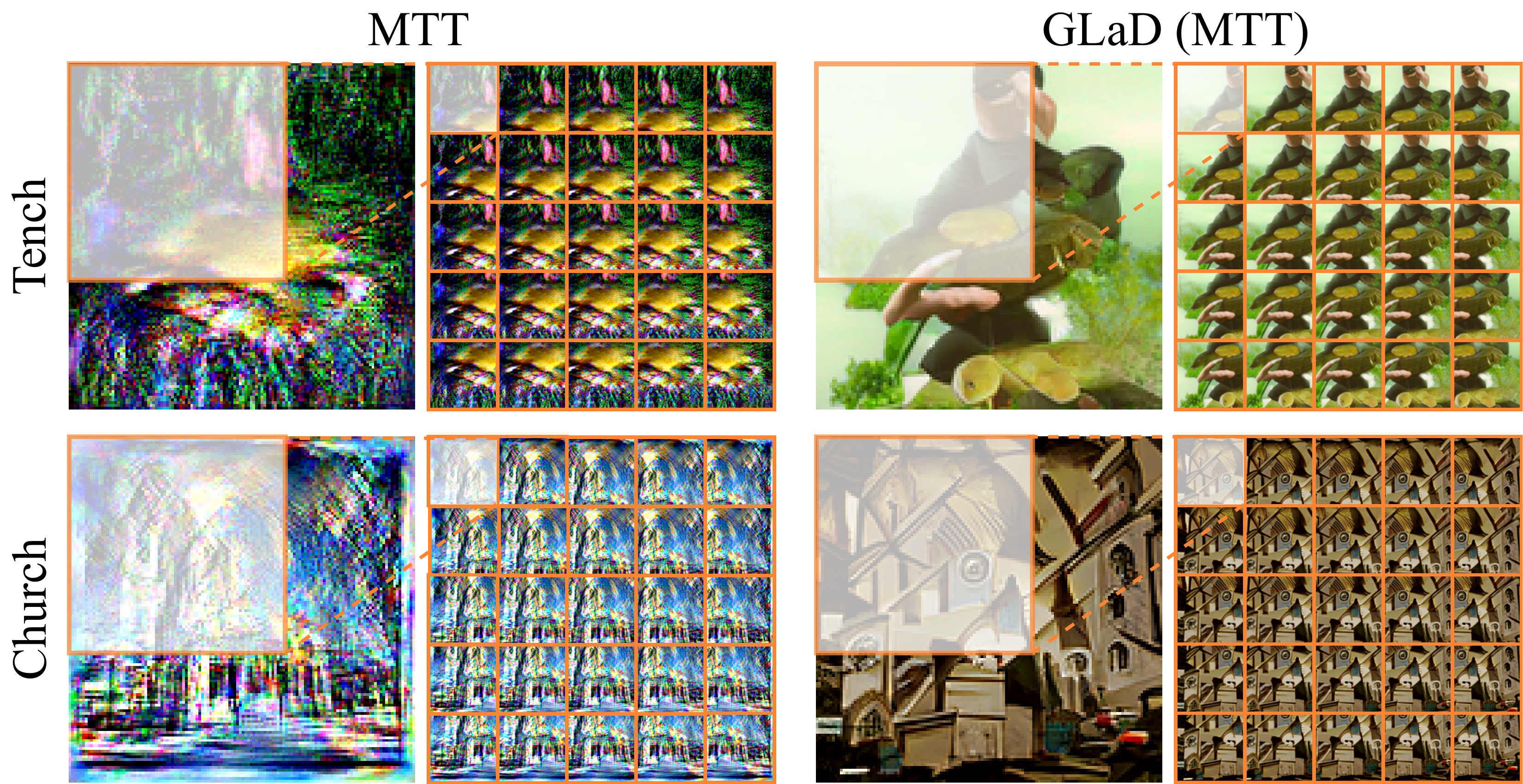}
  \vspace{-0.2in}
  \caption{\textbf{The result of sub-sampling of MTT and GLaD.} Visualization of the sub-sampling results for the Tench and Church classes from the Imagenette dataset, distilled using the MTT and GLaD methods. For each sample, the image on the left is the original distilled image, and the images on the right are the sub-images after sub-sampling. The original images selected are the first index images from each class.}
  \label{fig:sup_subsample}
  \vspace{-0.1in}
\end{figure}
Fig.~\ref{fig:sup_subsample} demonstrates examples of the results after applying sub-sampling to the distilled dataset. After distilling the Imagenette dataset using the MTT and GLaD methods, the images from the Tench and Church classes were extracted, and these are the original images shown on the left of each sample. Sub-sampling is then performed with hyperparameters set to $N$ = 5 and $R$ = 62.5\%, starting from the top-left corner of the original image. As a result, 25 sub-images are generated for each original image, which are displayed on the right of each sample.

\section{Future Works}
We aim to quantize the LADD to reduce storage requirements and improve training efficiency. Furthermore, we plan to explore the application of LADD in tasks that require higher computational costs, such as vision-language models. We will optimize the balance between dense and hard labels through ablation studies or by learning a weight parameter. Additionally, we intend to experiment with alternative static sub-sampling methods to enhance overall performance and scalability across diverse tasks.



\end{document}